\documentclass[lettersize,journal]{IEEEtran}
\ifCLASSINFOpdf
\else
\fi
\usepackage{algorithmic}
\usepackage{array}
\usepackage{graphicx}
\usepackage{cite}
\usepackage{amsmath}
\usepackage{amssymb}
\usepackage{booktabs}
\usepackage{multirow}
\usepackage{bbm}
\usepackage{color}
\usepackage{bm}
\usepackage{epsfig}
\usepackage{multicol}
\usepackage{booktabs}
\usepackage{stfloats}
\usepackage{makecell,rotating}
\usepackage{fancyvrb}
\usepackage[breaklinks,colorlinks]{hyperref}

\begin{document}

\title{Panoptic Perception: A Novel Task and Fine-grained Dataset for Universal Remote Sensing Image Interpretation}

\author{Danpei Zhao,~\IEEEmembership{Member,~IEEE,}
        Bo Yuan, Ziqiang Chen, Tian Li, Zhuoran Liu, Wentao Li, and Yue Gao 
\thanks{This work was supported by the National Natural Science Foundation of China under Grant 62271018. (Corresponding author: Danpei Zhao.)}
\thanks{Danpei Zhao, Bo Yuan, Tian Li are with the Image Processing Center, School of Astronautics,  Beihang University, Beijing 100191, China, and also with Tianmushan Laboratory, Hangzhou 311115, China (e-mail: zhaodanpei@buaa.edu.cn, yuanbobuaa@buaa.edu.cn, lit@buaa.edu.cn).}
\thanks{Ziqiang Chen, Zhuoran Liu, Wentao Li  and Yue Gao are with the Image Processing Center, School of Astronautics, Beihang University, Beijing 100191, China (e-mail: chenziqiang@buaa.edu.cn, liuzhuoran@buaa.edu.cn, canoe@buaa.edu.cn, bjlguniversity@163.com).}}

\markboth{Journal of \LaTeX\ Class Files,~Vol.~X, No.~X, X~X}%
{Shell \MakeLowercase{\textit{et al.}}}
%



\maketitle

\begin{abstract}
Current remote-sensing interpretation models often focus on a single task such as detection, segmentation, or caption. However, the task-specific designed models are unattainable to achieve the comprehensive multi-level interpretation of images. The field also lacks support for multi-task joint interpretation datasets. In this paper, we propose Panoptic Perception, a novel task and a new fine-grained dataset (FineGrip) to achieve a more thorough and universal interpretation for RSIs. The new task, 1) integrates pixel-level, instance-level, and image-level information for universal image perception, 2) captures image information from coarse to fine granularity, achieving deeper scene understanding and description, and 3) enables various independent tasks to complement and enhance each other through multi-task learning. By emphasizing multi-task interactions and the consistency of perception results, this task enables the simultaneous processing of fine-grained foreground instance segmentation, background semantic segmentation, and global fine-grained image captioning. Concretely, the FineGrip dataset includes 2,649 remote sensing images, 12,054 fine-grained instance segmentation masks belonging to 20 foreground things categories, 7,599 background semantic masks for 5 stuff classes and 13,245 captioning sentences. Furthermore, we propose a joint optimization-based panoptic perception model. Experimental results on FineGrip demonstrate the feasibility of the panoptic perception task and the beneficial effect of multi-task joint optimization on individual tasks. The dataset will be publicly available~\footnote{https://ybio.github.io/FineGrip/}.
\end{abstract}


\IEEEpeerreviewmaketitle

\section{Introduction}

\IEEEPARstart{R}{emote} sensing image (RSI) interpretation has witnessed a rapid development tendency in multiple tasks, including image classification~\cite{lu2007survey}, object detection~\cite{li2020object}, semantic segmentation~\cite{yuan2021review}, instance segmentation~\cite{su2019object}, image caption generation~\cite{lu2017exploring}, etc. However, these tasks only cover single task interpretation. Particularly, object detection focuses on locating and classifying individual objects within images. Instance segmentation identifies refined contours. Semantic segmentation aims to assign a class label to every pixel. While image caption generation attempts to provide a holistic understanding of RSIs. However, models of these tasks are usually designed independently and overlook the rich semantics and contextual relationships in RSIs. However, in the field of remote sensing observation, a very common and practical expectation is to achieve multi-level, fine-grained, perceptive interpretation of RSIs.

Recently, new efforts have emerged to promote a more comprehensive RSI interpretation. For example, panoptic segmentation~\cite{kirillov2019panoptic} combines the advantages of instance and semantic segmentation, enabling simultaneous foreground instance masking and background pixel classification. However, the datasets and research on RSI panoptic segmentation~\cite{de2022panoptic} are scarce. However, panoptic segmentation still focuses on pixel-level and instance-level interpretation. Another frontier research focus, i.e.,  fine-grained object recognition~\cite{sun2022fair1m} emerges as a pivotal task that focuses on identifying specific sub-categories of target objects. However, these tasks can not cope with multi-modal interpretation covering from pixel-level to image-level, lacking the comprehensive perception capacity and universal interpretation models across multi-modal tasks.

\begin{figure*}[htbp]
    \centering
    \includegraphics[scale=0.44]{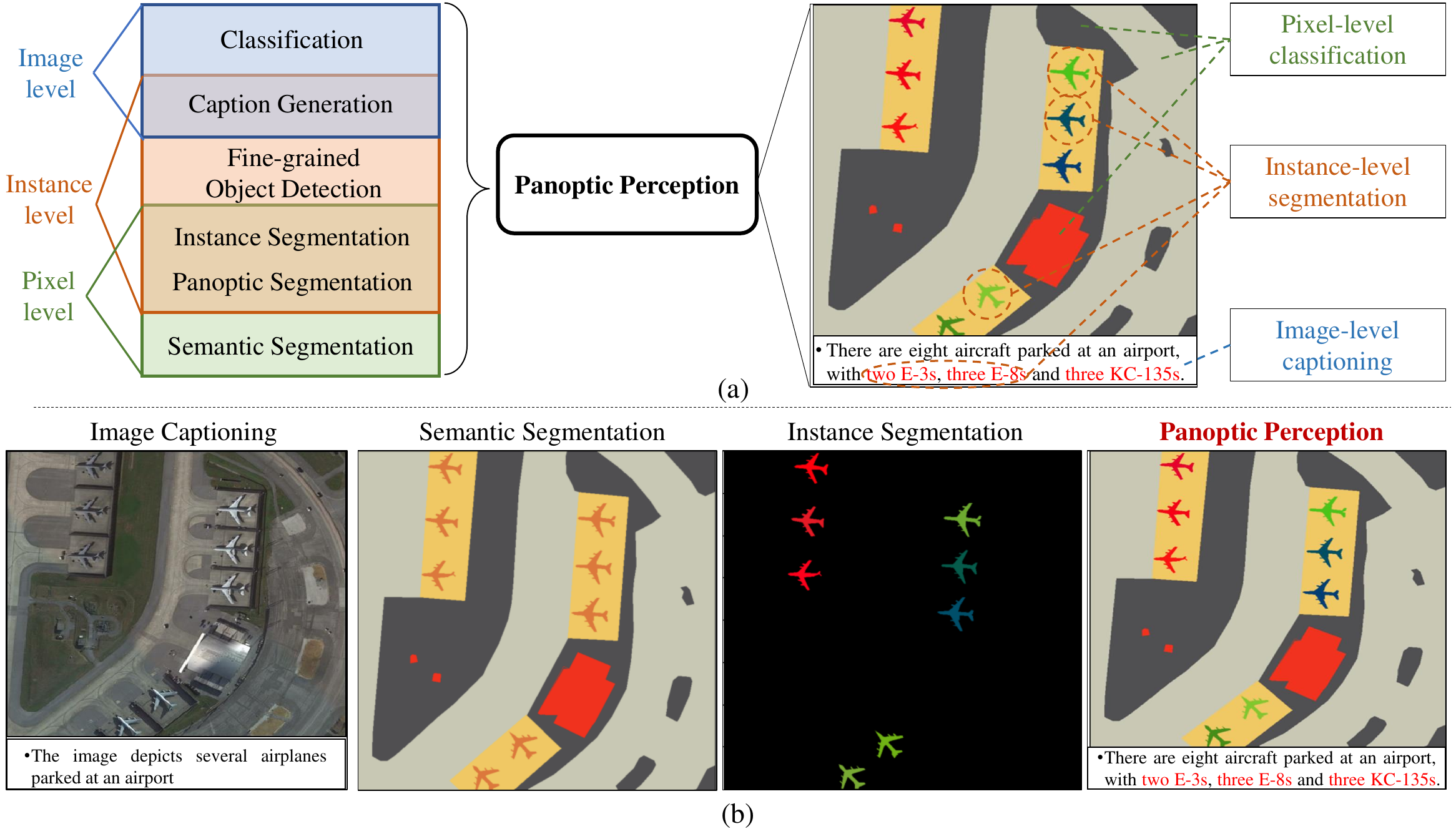}
    \caption{The proposed fine-grained Panoptic Perception task. (a) The definition and scope of panoptic perception Task. (b) Qualitative comparison of panoptic perception and other interpretation tasks.}
    \label{fig-motivation}
\end{figure*}

In this paper, we introduce Panoptic Perception, a novel task for comprehensive remote sensing image interpretation along with a new benchmark dataset FineGrip.  As shown in Fig.~\ref{fig-motivation}, panoptic perception can concurrently handle various sub-tasks across multi-level interpretation, including fine-grained instance segmentation of foreground instances, semantic segmentation for background areas, and image caption generation. This innovative task diverges from conventional tasks by not solely focusing on individual interpretation levels but leverages mutually reinforcing and interactive optimization.
The collaborative processing of multiple tasks necessitates that the model comprehensively understand the global contextual relationships and semantic information at different levels. This, in turn, amplifies the model capability to extract and utilize the abundant information in RSIs. The proposed panoptic perception integrates pixel-level, instance-level, and image-level understanding to construct a universal interpretation framework.

We construct a novel benchmark dataset named FineGrip to support the development of the new task. It comprises 2,649 remote sensing images, with fine-grained aircraft instance segmentation annotations, diverse background semantics and fine-grained sentence description annotations. To the best of our knowledge, this is the first dataset integrating fine-grained detection, instance segmentation, semantic segmentation, and fine-grained image caption annotations for RSIs. Additionally, we utilized the Segment Anything Model (SAM) to construct a semi-automatic segmentation annotation system. It fully leverages the robust zero-shot transfer capabilities of SAM, significantly enhancing the annotation efficiency of foreground segmentation.

To validate the feasibility of the proposed panoptic perception and the effectiveness of the dataset, we propose an end-to-end panoptic perception baseline model. The experimental results corroborated the viability of panoptic perception tasks and the beneficial impact of multi-task joint optimization on individual task enhancement.

The main contributions of this paper are summarized as follows:
\begin{itemize}
    \item We propose panoptic perception, a novel and comprehensive interpretation task that synchronously performs pixel-level, instance-level and image-level perception.
    \item We present FineGrip, which to our best knowledge, serves as the first fine-grained panoptic perception dataset to support the multi-task interpretation of RSIs.
    \item A new semi-automatic annotation system is designed to enhance annotation efficiency by combining the zero-shot generalization ability of foundation models with the fitting capability of supervised models.
    \item We introduce an end-to-end panoptic perception baseline method that offers referential research directions for subsequent studies.
\end{itemize}

\section{Related Works}
\subsection{Remote Sensing Datasets for Various Interpretation Tasks}
\indent Currently, a variety of task settings are employed in the interpretation of RSIs. These tasks range from pixel-level, instance-level to image-level, and low-level details to high-level semantics, providing a multi-faceted understanding of RSIs.

Semantic segmentation is a pixel-level interpretation task that aims to assign a category label to each pixel. There are multiple RSI semantic segmentation datasets constructed for various applications such as water area extraction~\cite{fernandes2020water}, cloud and fog detection~\cite{38cloud1, roynard2018paris}, land cover and vegetation classification~\cite{yang2010bag, alemohammad2020landcovernet, wang2021loveda, xia2017aid}, road segmentation~\cite{liu2018roadnet, van2018spacenet}, and disaster monitoring~\cite{rahnemoonfar2021floodnet}, etc. Semantic segmentation has achieved admirable accomplishments in these scenarios. However, the inability to distinguish between foreground instances limits its effectiveness in tasks focusing on specific foreground objects.

Instance-level interpretation tasks include object detection and instance segmentation. The former aims to locate and identify objects of interest. There are numerous datasets constructed for detecting typical objects such as airplanes, ships, oil tanks, bridges, and vehicles~\cite{xia2018dota, li2020object, lam2018xview, zou2017random}. The emerging fine-grained object detection tasks~\cite{sun2022fair1m, sun2022scan} require recognizing subtle distinctions among sub-categories of objects. It goes beyond merely identifying a general category like an \textit{airplane} and extends to specifying the exact type, such as \textit{Boeing 747}. Instance segmentation~\cite{liu2017high, waqas2019isaid} provides an additional pixel-level mask for each identified object, enriching the understanding of the exact contour of objects. The localization, identification, and segmentation of typical remote sensing targets are crucial in applications like disaster loss assessment~\cite{lv2022spatial} and military surveillance~\cite{li2018hsf}. However, a challenge emerges from the sparse distribution of foreground objects in RSIs. Relying solely on the foreground object features could result in the significant neglect of invaluable background information.

The goal of image caption generation for RSIs is to perceive the overall situation and automatically generate text that describes the content of the images.  In contrast to the prevalent focus on delineating actions and behaviors in natural images, datasets for RSI captioning~\cite{xia2017aid, lu2017exploring, chen2020spatial} prioritize the recognition of surface attributes, geographical formations, and the interrelationships among geophysical entities, etc. It provides a concise overview of the image content. However, its inherent generality and ambiguity may cause models to overlook critical details such as the exact number, size, and type of targets.

Despite the specific significant progress, these tasks remain focused on specific scenarios and fail to fully extract and utilize the wealth of rich information contained in RSIs. A more thorough understanding of RSIs necessitates multi-modal and multi-task interpretation.

\subsection{Panoptic Segmentation for RSIs}
\indent Panoptic segmentation~\cite{kirillov2019panoptic} is a generalized segmentation task that offers a unified framework to address background semantic segmentation (Stuff) and foreground instance segmentation(Things). Specifically, it aims to assign a category label to every pixel in an image and distinguish different instances of the same category. Standard benchmark datasets for panoptic segmentation tasks include COCO-Panoptic~\cite{lin2014microsoft} and Cityscapes~\cite{cordts2016cityscapes}. However, panoptic segmentation in RSIs faces the following challenges. First, there is a lack of high-quality fine-grained datasets. PASTIS~\cite{garnot2021panoptic} proposes an open-source dataset for panoptic segmentation of RSIs. However, the restricted image size of 128$\times$128 limits its application. BSB aerial dataset~\cite{de2022panoptic} contains fine-grained instance annotations only for building targets. On the other hand, currently most methods are carried out on a few categories with high differentiation such as~\cite{RSDINO, 10105645}, and there is a lack of fine-grained labels with foreground instances. In addition, current panoptic segmentation methods lack the ability of multi-modal interpretation.

In this paper, we extend RSI interpretation from single-task to multi-modal panoptic perception. A new panoptic perception dataset is proposed, which supports multi-modal tasks including pixel-level classification, instance-level segmentation and image captioning.

\subsection{Efficient Data Annotation}
As the foundation for training robust segmentation models, accurate annotations offer benchmarks against model predictions and guide optimization. Over the years, high-quality annotated datasets have notably advanced RSI segmentation.

Traditional manual annotation tools, such as Labelme \cite{russell2008labelme}, achieve high precision by allowing users to delineate target contours with polygons. When it comes to annotating complex-shaped targets, these tools prove labor-intensive. With the expanding volume of RSI datasets, Improving annotation efficiency becomes a challenge. 

The zero-shot learning capabilities of foundation models make them highly effective for annotating massive datasets. For example, the Segment Anything Model (SAM)~\cite{kirillov2023segment}, with its ability to effectively segment specific areas based on provided box prompts, inspires the development of numerous semi-automatic annotation systems. ISAT~\cite{ISAT} integrates Labelme with SAM, facilitating interactive semi-automatic segmentation annotation. It enables users to craft and modify boundaries, which are taken as prompts into SAM to achieve target-specific segmentation. Semantic SAM~\cite{chen2023semantic} emerges as the first open framework employing SAM for semantic segmentation tasks. It smoothly enhances performance without fine-tuning the weights of SAM. Grounded-Segment-Anything fuses SAM with Grounding-DINO~\cite{liu2023grounding}, delivering zero-shot detection results through textual input. The output bounding boxes then serve as prompts for instance segmentation.

The lack of familiarity with typical remote sensing targets poses challenges for the use of SAM in RSI annotation. While SAMRS~\cite{wang2023samrs} effectively converts bounding boxes into masks using original box annotations as prompts, It constrains category expansion because of the reliance on existing detection annotations. Meanwhile, fine-tuning SAM can address domain discrepancies but needs a large amount of training data and incurs significant computational costs. We aim to leverage SAM for quality annotations with limited resources and fewer images. Thus, instead of fine-tuning its parameters, we use SAM for inference and employ its results to train a more efficient segmentation model for enhanced outcomes.

\begin{figure*}[htbp]
    \centering
    \includegraphics[scale=0.42]{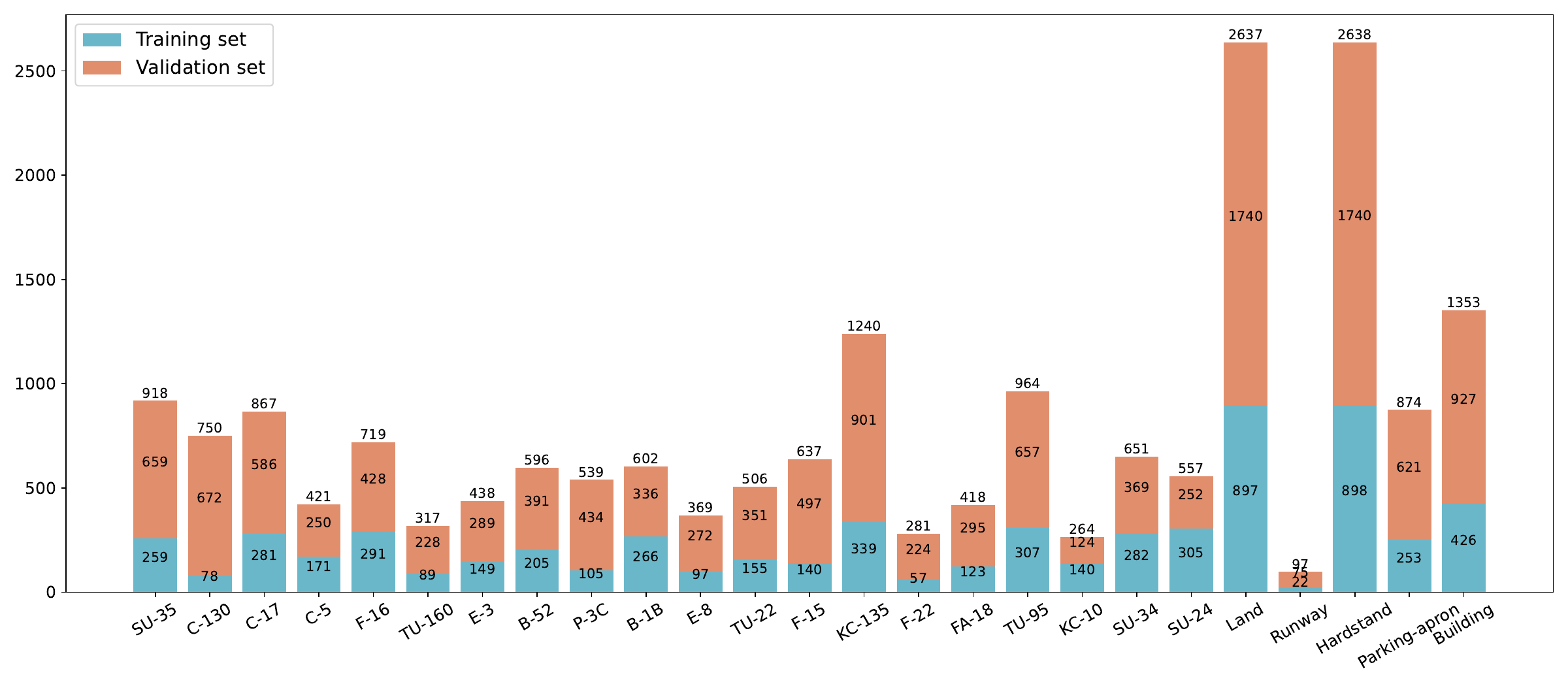}
    \caption{Number of per-category masks in the FineGrip dataset across training and validation sets.}
    \label{fig-dataset_info}
\end{figure*}

\section{Task Definition}
In this paper, we propose a fine-grained, unified framework to simultaneously achieve pixel-level, instance-level, and image-level interpretation of RSIs. As illustrated in Fig.~\ref{fig-motivation}, beyond, our proposed task exceeds traditional single-task and necessitates models to extract more comprehensive contextual features and enables the joint interpretation of multiple tasks at various levels:

1) At the image level, the task demands the model to generate a concise overview of the entire content of the image and output this overview using natural language. 

2) At the instance level, the model identifies the fine-grained categories of all foreground objects, distinguishes between different instances within the same category, and predicts an accurate contour for each instance. The task also demands the model to specify the number and specific categories of all foreground instances in its descriptive sentences.

3) At the pixel level, the task necessitates that each pixel in the image be assigned a distinct category of either foreground or background. Moreover, pixels associated with different foreground instances must be assigned a unique identifier.

Given an image $I\in \mathbb{R}^{H \times W \times 3}$, we define a set of words $\textit{Wds}=\{wd_1,wd_2,...,wd_W\}$ and a set of categories $C^P =\{c_1,c_2,...,c_C\}$, where W, C is the total number of words and categories, respectively. $C^P$ can be further divided into a foreground category (things) set $C^{Th}$ and a background category (stuff) set $C^{St}$, where $C^{Th} \cap C^{St} = \emptyset$. The objectives of the fine-grained panoptic perception task are 
 formally defined as follows:

1) For any given pixel $(x,y)$ in the image, the model is required to predict both the category and the instance id of the pixel, denoted as $(c_{x,y}, id_{x,y})$. All pixels within the same instance should share identical category and number identifications. When a pixel belongs to a stuff category, the predicted instance id should be $\emptyset$.

2) Considering a maximum sentence length $L$, the model should generate a descriptive sentence for the image, represented as $\{w_1,w_2,...,w_L|w_i \in \textit{Wds}\}$. This sentence must include information about the number and types of foreground objects in the image.

Fine-grained panoptic perception demands the consistency of perception results across sub-tasks. As shown in Fig.~\ref{fig-motivation}(a), the caption concerning the quantities and types of foreground instances should align with the segmentation results.

Unlike a simple combination of multiple independent tasks, fine-grained panoptic perception emphasizes the complementarity and joint optimization among sub-tasks to achieve integrated image perception. Semantic segmentation provides detailed information about the background areas of RSIs, which aids in detecting and masking foreground instances. Meanwhile, fine-grained instance segmentation of foreground classes yields information on the number, category, and relative position of targets, which serve as critical materials for fine-grained caption generation. Moreover, accurate descriptive sentences can also help achieve more precise segmentation due to the constraint of consistency in perception results.

For the segmentation sub-task, following~\cite{kirillov2019panoptic}, we employ Panoptic Quality (PQ) to assess the performance and utilize $PQ^{th}$ and $PQ^{st}$ to measure the segmentation quality on \textit{things} and \textit{stuff} classes. Moreover, Recognition Quality (RQ) and Segmentation Quality (SQ) are applied to analyze the recognition and segmentation performance. As for image caption generation, we use the BLEU~\cite{papineni2002bleu} metric to examine the overall description quality. Details of these evaluation metrics are introduced in section V.

\begin{figure*}[htbp]
    \centering
    \includegraphics[scale=0.5]{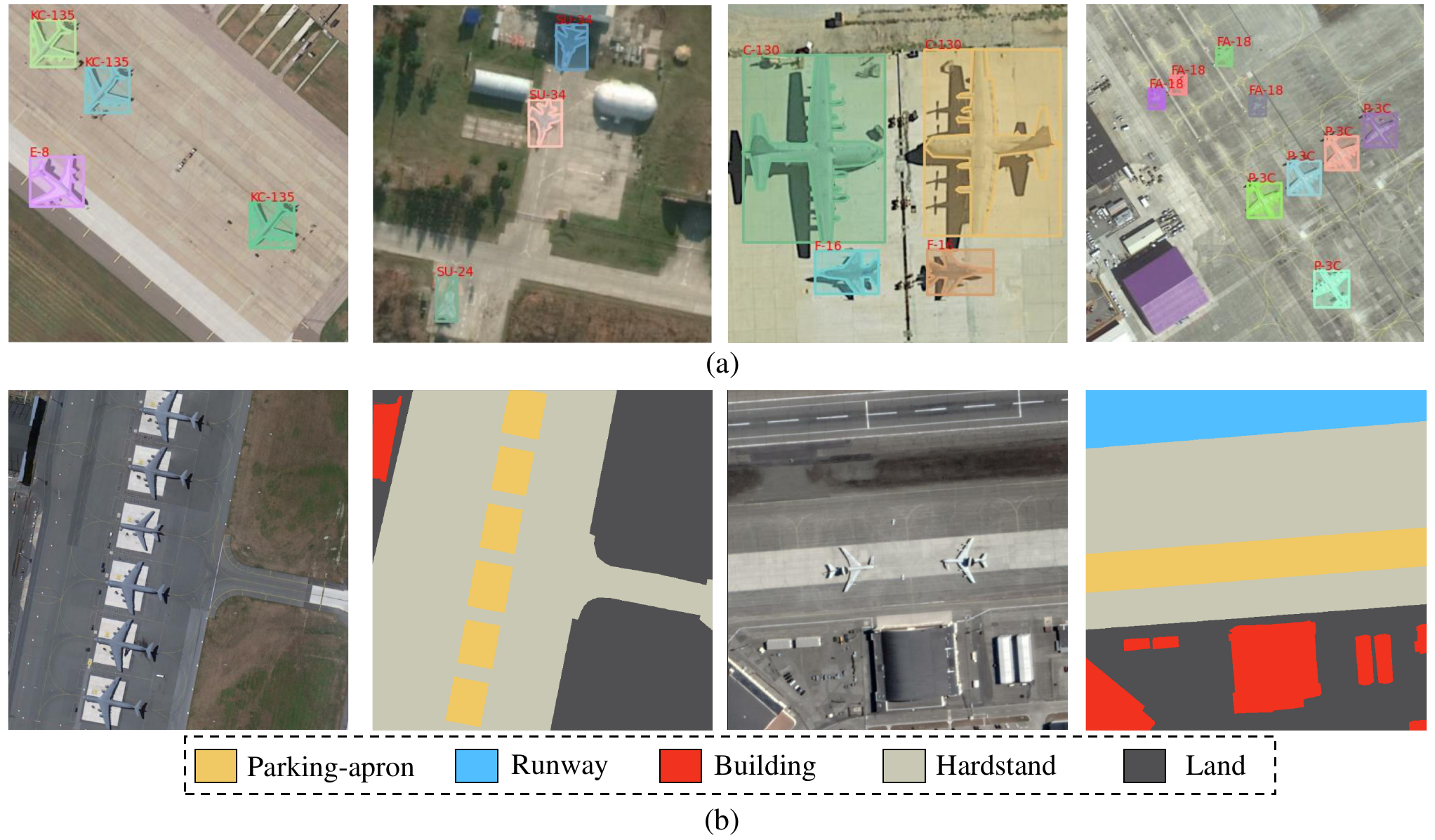}
    \caption{Examples of Annotations in FineGrip. (a) Foreground fine-grained instance segmentation annotations.(b) Background stuff semantic segmentation annotations.}
    \label{fig-FineGrip_example}
\end{figure*}

\begin{figure*}[htbp]
    \centering
    \includegraphics[scale=0.5]{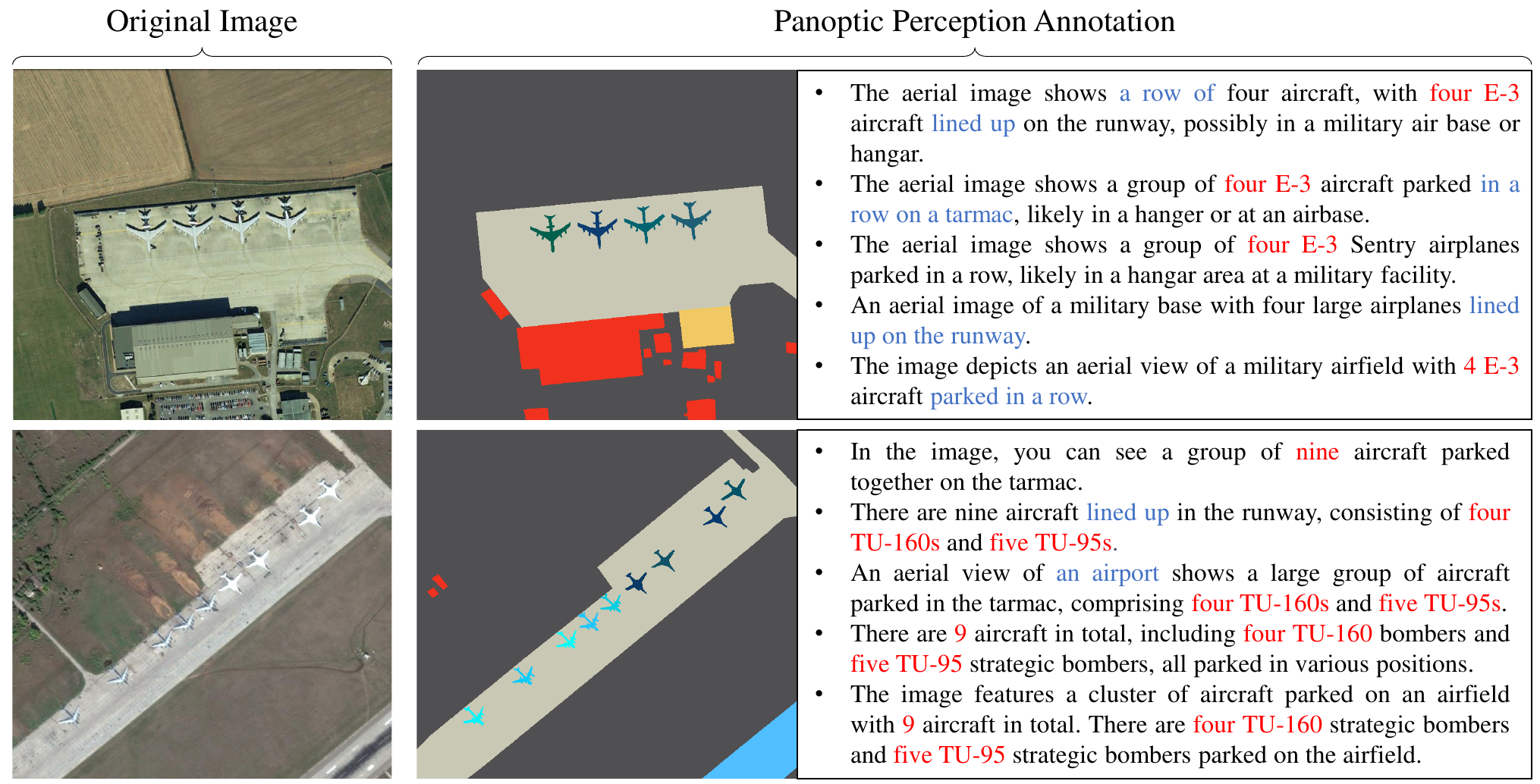} 
    \caption{Examples of Per-Image Annotations from FineGrip. In the captions, red highlights detail the fine-grained categories and quantities of aircraft instances, while blue highlights indicate relationships between instances.}
    \label{fig-FineGrip_annotation}
\end{figure*}

\section{Dataset Construction}
Existing datasets originate from diverse sources and are usually annotated for specific tasks, making them inadequate for the unified perception task defined in this paper. Therefore, we develop a fine-grained panoptic perception benchmark dataset based on a novel semi-automatic annotation system. In this section, we will provide a detailed introduction to the benchmark dataset and the semi-automatic segmentation annotation system.

\begin{figure*}[htbp]
    \centering
    \includegraphics[scale=0.43]{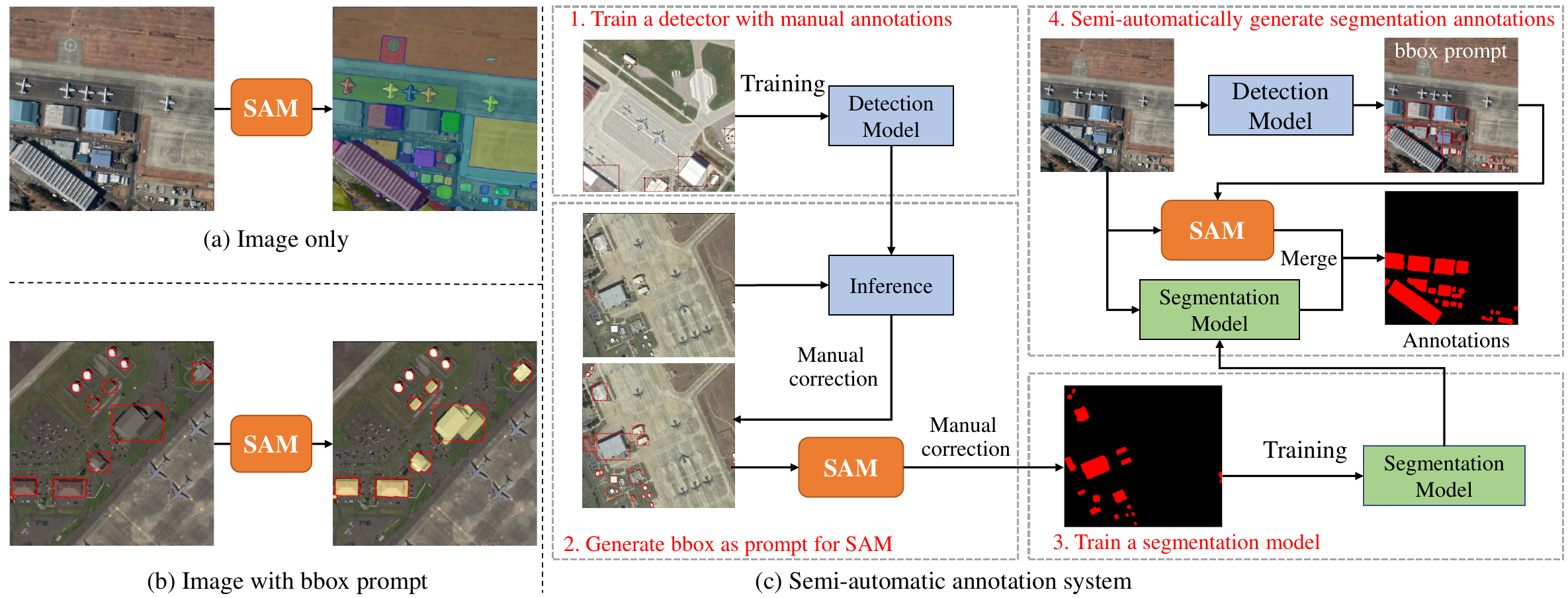}  
    \caption{Examples of Per-Image Complete Annotations from FineGrip. In the captions, red highlights detail the fine-grained categories and quantities of aircraft instances, while blue highlights indicate relationships between instances.}
    \label{fig-SAAS} 
\end{figure*}

\subsection{The FineGrip Dataset}
To facilitate the research on the new task, we constructed a Fine-Grained Panoptic Perception (FineGrip) Dataset encompassing 20 fine-grained foreground aircraft categories and 5 background stuff categories. The sample images in the proposed FineGrip are mainly derived from MAR20~\cite{mar20} but with further cleaning and replenishment. As seen in Table~\ref{table-dataset}, the original MAR20 only contains bounding box annotations. However, we extend the annotations to support fine-grained instance segmentation, semantic segmentation, and image caption generation tasks. Concretely, FineGrip comprises 2,649 remote sensing images, 12,054 instance segmentation masks across 20 \textit{things} categories, and 7,599 background semantic masks across 5 \textit{stuff} categories. And 13,245 sentences with fine-grained category indications. The \textit{things} categories include SU-35, C-130, C-17, C-5, F-16, TU-160, E-3, B-52, P-3C, B-1B, E-8, TU-22, F-15, KC-135, F-22, FA-18, TU-95, KC-10, SU-34, and SU-24. For representative convenience, the categories are respectively represented by $A1\sim A20$. While the background \textit{stuff} categories comprise \textit{Land}, \textit{Runway}, \textit{Hardstand}, \textit{Parking-apron} and \textit{Building}. Fig.~\ref{fig-dataset_info} illustrates the number of segmentation masks for each category in the training/testing sets. The dataset contains 901 images for training and 1,748 images for testing.

 \begin{table}[tbp]
 	\caption{The proposed FineGrip dataset.}
 	\centering
 	\setlength{\tabcolsep}{0.8mm}{
 		\begin{tabular}{l|ccccccc}
 			\toprule[0.4mm]
                 Dataset &\makecell{Class\\num.} &\makecell{Sample\\num.}&Bbox& Masks &Caption &\makecell{Fine\\grained} & \makecell{Cross\\modal}\\
                \midrule
                 MAR20~\cite{mar20}&20&3842&$\checkmark$&$\times$&$\times$&$\checkmark$ &$\times$ \\
                 BSB~\cite{de2022panoptic} &15&3400&$\checkmark$&$\checkmark$&$\times$&$\times$ &$\times$ \\
                \midrule
                 \textbf{FineGrip} &25&2649&$\checkmark$&$\checkmark$&$\checkmark$&$\checkmark$ &$\checkmark$ \\
 			\bottomrule[0.4mm]
 	\end{tabular}}
 	\label{table-dataset}
 \end{table}

The FineGrip dataset focuses on airport scenes, where various aircraft models are our primary foreground categories of interest. We annotated 20 fine-grained aircraft targets with instance-specific segmentation. Fig.~\ref{fig-FineGrip_example} showcases some examples of FineGrip.
In the context of background (\textit{stuff}) categories, we prioritize areas closely related to aircraft targets. We define \textit{Runway} as a long straight road marked with lines. A \textit{Parking-apron} is a notable region where aircraft are parked for extended periods. \textit{Hardstand} refers to areas, other than the previously mentioned two kinds, where aircraft can taxi. Buildings are categorized as background rather than foreground because we are not concerned with the specific instance segmentation of buildings in this scene. We merely need to identify which areas in the image pertain to buildings. Moreover, unlike aircraft targets, buildings often present a closely interconnected structure, complicating instance-level differentiation. While \textit{Land} designates areas in the image other than the foreground aircraft targets and the mentioned stuff categories. Fig.~\ref{fig-FineGrip_example} shows some examples of segmentation annotations for \textit{things} and \textit{stuff} categories.

As for the fine-grained image caption task, we place emphasis on the precise number and models of the foreground targets, as well as the relationships among the physical features. We generate five captions from five different annotators for each image to foster caption diversity. Ultimately, by integrating fine-grained instance segmentation, background semantic segmentation, and fine-grained caption annotations, we establish the FineGrip dataset. Fig.~\ref{fig-FineGrip_annotation} displays some complete annotation examples from FineGrip.

Compared to conventional interpretation tasks and the recently proposed RSI panoptic segmentation dataset, our FineGrip exhibits significant characteristics in the following aspects:

\begin{itemize}
    \item \textbf{\textit{Abundant fine-grained semantic categories.}} The proposed FineGrip covers 20 fine-grained foreground categories and 5 densely labeled background categories. The samples from different categories possess diverse semantics, wide terrain scenes and complex semantic relations, etc. Besides, it also meets the practical challenges of small inter-class differences and large intra-class differences.
    \item \textbf{\textit{Broader granularity of caption sentences.}} The captioning annotations span from general to specific granularity, delivering a comprehensive view of images. It is also fine-grained and consistent with the pixel-level annotations. In addition, the complex semantic relations are depicted in detail to achieve human-like perception from a global perspective.
    It gives a general overview of the image and pinpoints the exact count and model of primary targets. 
    \item \textbf{\textit{Affinity exploration of foreground-background relationships.}} In FineGrip, the foreground and background categories share a close relationship. For example, aircraft is predominantly parked in \textit{Parking apron} or \textit{Hardstand} areas, but rarely seen in \textit{Land} areas. Moreover, building zones are often separated by hardstand areas. Such objective factors suggest that the panoptic perception model should consider these semantic relations, i.e., the foreground recognition and the background segmentation have the potential of mutual boosting.
    \item \textbf{\textit{Synergized multi-tasking.}} Coordinating the instance segmentation and image caption tasks, which both identify the target quantity and sub-category, can mutually enhance their performance.
    
\end{itemize}
\begin{figure*}[htbp]
    \centering
    \includegraphics[scale=0.32]{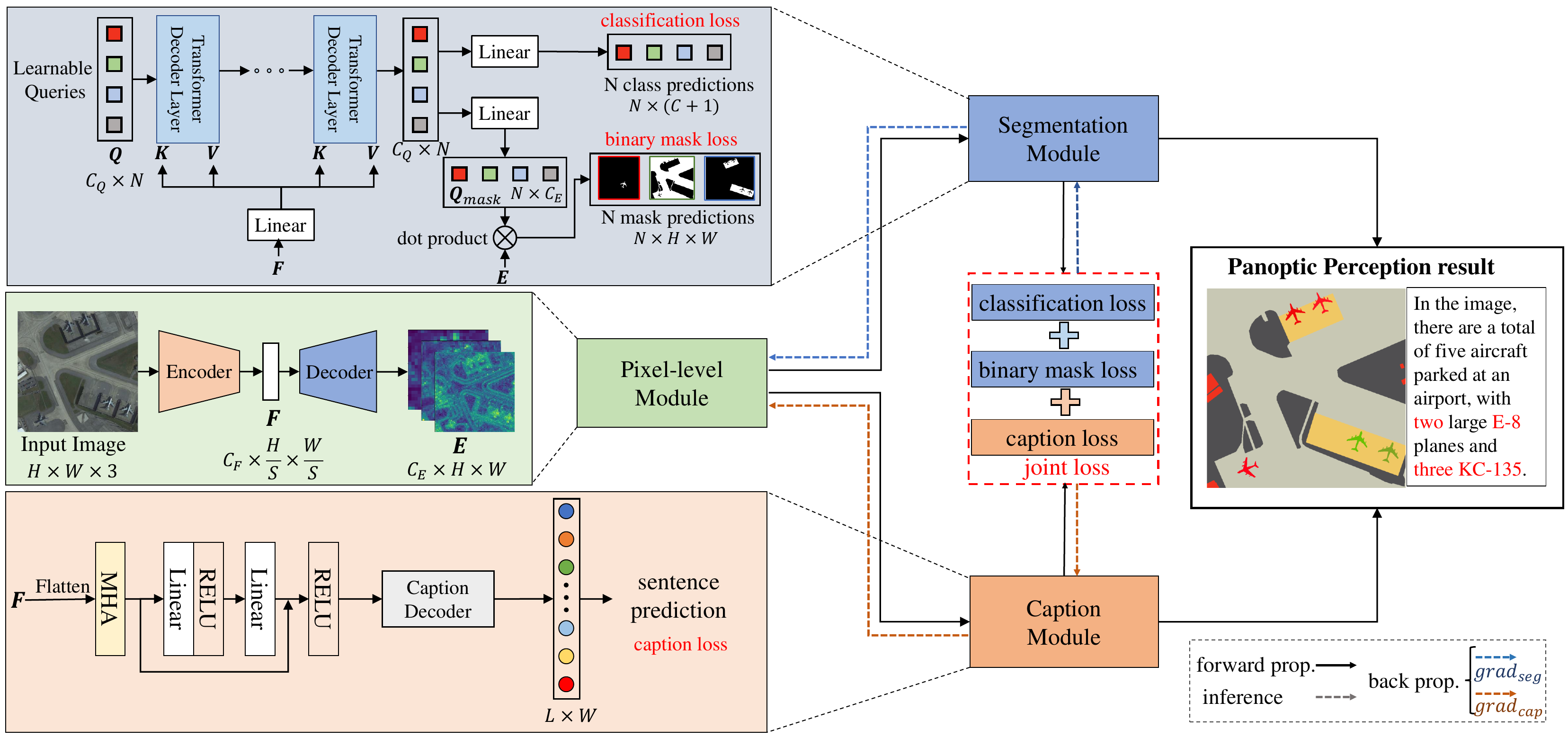} 
    \caption{Overall architecture of our proposed panoptic perception model. \textcolor{blue}{$grad_{seg}$} and \textcolor{red}{$grad_{cap}$} indicate the gradient calculated from the segmentation loss and the caption loss. }
    \label{fig-network} 
\end{figure*}
\subsection{Semi-Automatic Annotation System}

Due to its robust generalization capabilities, SAM is widely applied in the segmentation of natural images. When provided with high-quality prompts, annotation systems based on SAM can generate high-quality unlabeled segmentation masks. Moreover, fine-tuning SAM with a small amount of annotated data can yield good performance in various downstream segmentation tasks.

\indent However, SAM does not achieve comparable results for RSIs. As shown in Fig.~\ref{fig-SAAS}, there are primarily two approaches using SAM for RSI segmentation annotations: directly inputting the image (a) and utilizing manually annotated bounding boxes as prompts (b). However, method (a) struggles with the substantial domain difference between natural images and RSIs. Meanwhile, method (b) does not eliminate the manual effort required for bounding box annotations.

\indent The key to efficiently annotating RSIs with SAM is fully leveraging its zero-shot segmentation capabilities while compensating for its lack of RSI-specific knowledge. To address these challenges and improve annotation efficiency, we design a novel semi-automatic annotation system based on SAM, supplemented by a supervised detection and segmentation model, as shown in Fig.~\ref{fig-SAAS}(c). 

In the beginning, we manually annotate bounding boxes for a small set of images to train a detector. To ensure the quality of annotations for unseen images, bounding boxes generated by the detection model undergo manual checks. Subsequently, the predicted box results serve as prompts and are fed into SAM to segment certain targets in the image. We will train a supervised segmentation model after refining these segmentation results.

Note that the described process is iterative, which means the detection and segmentation results from the current step directly feed into the training data for the next step.

To annotate unseen images, we first employ the detection model to obtain box prompts. Then, both SAM and the trained segmentation model are used to predict segmentation results. We achieve the final segmentation annotations by merging results from SAM and our trained model. In practice, simply obtaining intersection areas can effectively combine the segmentation results.

\section{Joint Optimization-Based Panoptic Perception Method}
As shown in Fig.~\ref{fig-network}, we propose a simple baseline method to perform end-to-end fine-grained panoptic perception. It achieves multi-task, multi-stage information interaction by optimizing a joint loss function. The model consists of three parts: the pixel-level module, the panoptic segmentation module, and the image caption module. They are responsible for encoding image features, predicting object masks, and generating captions, respectively. The features derived from the pixel-level module are utilized for both mask prediction and caption generation. Therefore, during optimization, the gradient calculations from the loss functions of both segmentation and captioning tasks will contribute to updating the pixel-level module parameters.
\subsection{Pixel-level Module}

The pixel-level module primarily consists of an image encoder and a decoder. Consider an image of size $H \times W$. The image encoder produces a down-sampled feature $F$ of size $C_F \times \frac{H}{S} \times \frac{W}{S}$, where $S$ is the output stride and $C_F$ is the number of channels of the encoded feature. The decoder then upsamples $F$ to obtain per-pixel feature embeddings $E \in \mathbb{R}^{C_E \times H \times W}$, which are used for subsequent mask predictions.  We apply ResNet-50 \cite{he2016deep} as the image encoder and Transformer Encoder \cite{vaswani2017attention} with convolutional layers as the image decoder. Specifically, we feed the flattened $F$ into the Transformer decoder, reshape the obtained embeddings back to $hxw$, and then use deconvolution layers for upsampling. By default, we set $S=256$, $C_F=256$, and $C_E=256$.

\subsection{Segmentation Module}
We regard both instance and semantic segmentation as mask classification problems and handle them with a Transformer-based architecture. Initially, $N$ learnable queries $Q \in \mathbb{R}^{C_Q \times N}$ are initialized, where $C_Q$ is the dimension of the queries. The features $F$ obtained from the pixel-level module are used as keys ($K$) and values ($V$). We use a standard Transformer decoder to iteratively update $Q$. Similar to DETR, we will save the results of each decoder layer. 

A typical Transformer decoder layer computation consists of three parts: self-attention on $Q$, cross-attention between $Q$, $K$, and $V$, and a feed-forward neural network. Readers are referred to \cite{vaswani2017attention} for more details. Note that we did not employ masked attention as there is no temporal relationship among the queries.

Through interactions with other queries and the image encoding features, the query can learn the features of various targets and their positional information within the image. Subsequently, we use these information-rich queries for mask classification and generation.

In the mask classification branch, the encoded query undergoes a linear transformation to yield a classification result of $N \times (C + 1)$, where $C$ is the total number of foreground and background categories. An additional category $\emptyset$ represents \textit{no object}.

In the mask generation branch, the query is projected into mask embeddings $\mathbf{Q}_{\text{mask}} \in \mathbb{R}^{N \times C_E}$, which has the same channel dimensions as the per-pixel feature embeddings. Subsequently, the dot product is performed between the $i$-th mask embedding and the matrix $\mathbf{E}$, followed by applying a sigmoid function to generate the $i$-th mask prediction result.

Referring to~\cite{MaskFormer}, we adopt the Hungarian matching \cite{carion2020end} to generate a one-to-one mapping between the mask prediction results and the ground truth. Consider the predictions 
\begin{equation}
    r^{\text{pred}} = \left\{ (p_i, m_i) \mid p_i \in \mathbb{R}^{C+1}, m_i \in [0,1]^{H \times W} \right\}_{i=1}^{N} 
\end{equation}

and the M Ground Truth (GT) masks
\begin{equation}
    r^{\text{gt}} = \left\{ (c_j^{gt}, m_j^{gt}) \mid c_j^{gt} \in \{1,2,...,C\}, m_j^{gt} \in [0,1]^{H \times W} \right\}_{j=1}^{M},
\end{equation}

We denote $i = \sigma(j)$ as the mapping from $r^{gt}_j$ to $r^{pred}_i$, and use 
\begin{equation}
    -p_i(c_j^{\text{gt}}) + \mathcal{L}_{mask}(m_i, m_j^{gt})
\end{equation}
as the edge weight in the bipartite graph for the Hungarian matching, where $\mathcal{L}_{mask}$ is the binary mask loss. Normally, we assume $N \gg M$. As a result, some prediction results will be matched to the \textit{no object} category $\emptyset$. We then pad the M GT masks to match the prediction size N using $\emptyset$.

giving a matching $\sigma$, the mask prediction loss consists of a cross-entropy loss and a binary mask loss for each $pred-gt$ pair:
\begin{equation}
	\begin{aligned}
\mathcal{L}_{seg}(r, r^{gt}) = & \sum_{j=1}^{N} [ -\log p_{\sigma(j)}(c_j^{gt}) +\\ & \mathbbm{1}_{c_j^{gt} \neq \emptyset} L_{mask}(m_{\sigma(j)}, m_j^{gt}) ]
	\end{aligned}
\end{equation}
Particularly, we apply Dice Loss \cite{sudre2017generalised} for the binary mask loss.
\begin{figure}
    \centering
    \includegraphics[scale=0.44]{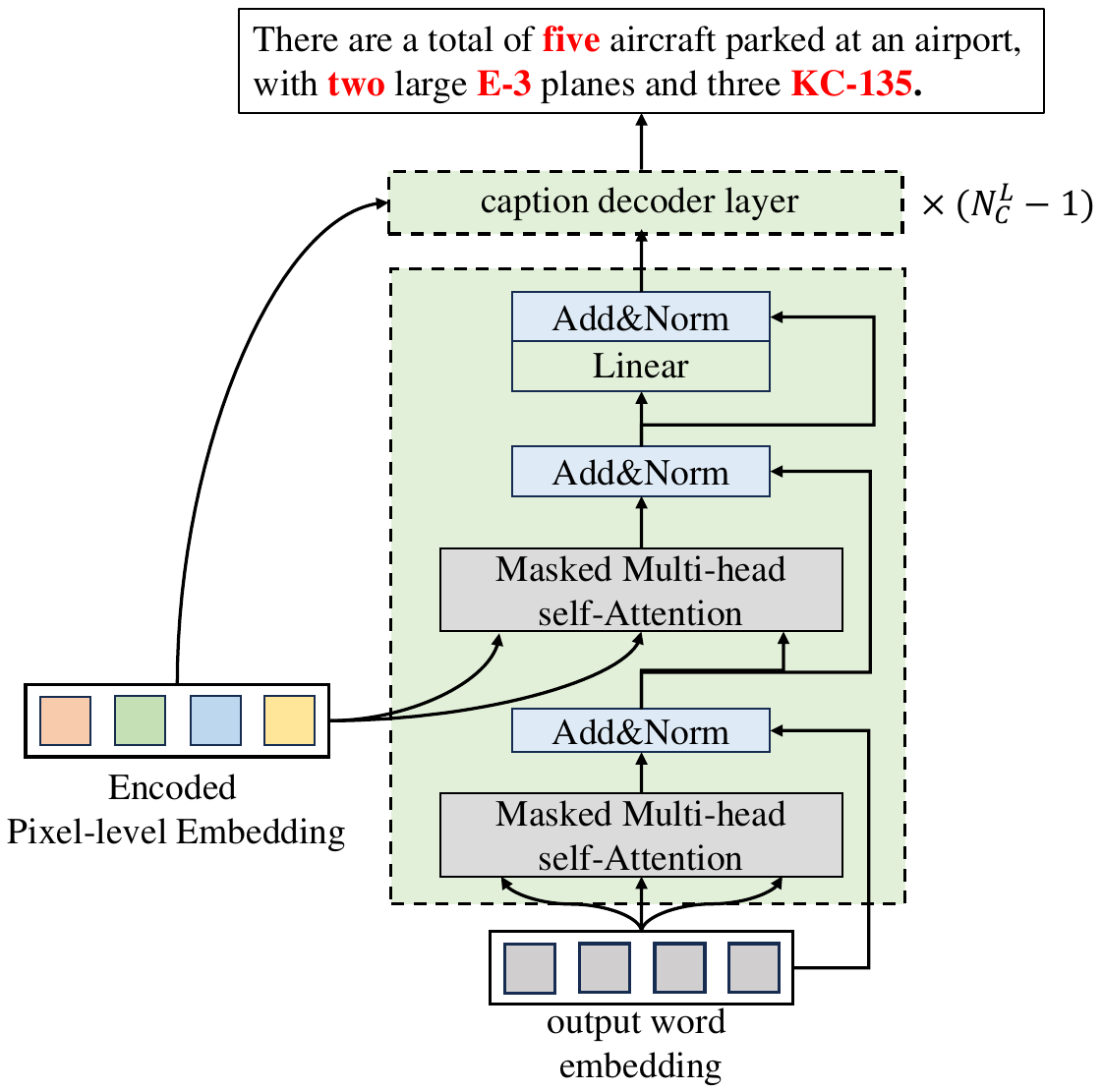}
    \caption{The model-agnostic task-interactive caption module.}
    \label{fig-caption_module}
\end{figure}

\subsection{Caption module}
To achieve end-to-end collaborative joint training, we propose a model-agnostic caption module that utilises a transformer decoder, which extracts more abstract semantic information from image features and outputs a sequence of words for description.
As depicted in Fig.~\ref{fig-network}, the encoded image features $F$ obtained from the Pixel-level module:
\begin{equation}
F' = \text{MHA}(F, F, F)
\end{equation}
\begin{equation}
\hat{F} = \text{Linear}(\text{ReLU}(\text{Linear}(F'))) + F'
\end{equation}
\begin{equation}
\hat{F} = \text{ReLU}(\hat{F})
\end{equation}
As depicted in Fig.~\ref{fig-caption_module}, we use a model-agnostic Transformer-based decoder to perform caption generation.

Following\cite{vaswani2017attention}, the Transformer-based-decoder used for sentence prediction consists of multiple decoder layers, each having a masked self-attention mechanism, a cross-attention mechanism, and a feed-forward neural network. We initialize a word representation $X_0 \in \mathbb{R}^{L \times C_F}$ with the START identifier as queries and use $T$ as keys and values for the cross-attention. Similarly, we use a linear classifier for word prediction and compute the loss function using cross-entropy. The number of the decoder layers is denoted as $N_{C}^{L}$.

The output embedding of the decoder is denoted as $h_t$, and the probability distribution over $W$ words is computed as:
\begin{equation}
    p_t = \text{Softmax}(\text{Linear}(h_t)),\ p_t \in [0,1]^{W}
\end{equation}
\indent Considering the word predictions $\{ p_t \}_{t=1}^L$ and the GT captions $\{ wd_t | wd_t \in \{1,2,..., W\} \}_{t=1}^L$, we compute the caption generation loss using cross-entropy as:

\begin{equation}
\mathcal{L}_{cap} = - \sum_{t=1}^{N_{C}^{L}} \log p_t(wd_t)
\end{equation}

\subsection{Overall Objective}
To jointly optimize segmentation and caption branch, the total loss function is the weighted sum of $\mathcal{L}_{seg}$ and $\mathcal{L}_{cap}$:
\begin{equation}
    \mathcal{L}_{total} = \mathcal{L}_{seg} + \lambda \mathcal{L}_{cap}
    \label{eqn-lambda}
\end{equation}
where $\lambda$ is the weight to control the contribution of the caption module.

\section{Experiments}
In this section, we validate the effectiveness of our proposed unified baseline in handling fine-grained panoptic segmentation tasks through experiments on the FineGrip dataset.

\subsection{Evaluation Metrics}
Referencing \cite{kirillov2019panoptic}, we utilize the Panoptic Quality (PQ) to uniformly measure the performance of instance segmentation and semantic segmentation:
\begin{equation}
\text{PQ} = \frac{\sum_{(p,g) \in \text{TP}} \text{IoU}(p, g)}{|\text{TP}| + \frac{1}{2}|\text{FP}| + \frac{1}{2}|\text{FN}|}
\end{equation}
where $\text{TP}$, $\text{FP}$, and $\text{FN}$ represent true positives, false positives, and false negatives, respectively. A pair $(p,g) \in \text{TP}$ denotes a match between the predicted mask and the ground truth. Further, the PQ can be decomposed into Segmentation Quality (SQ) and Recognition Quality (RQ) to separately measure the performance of segmentation and recognition:
\begin{equation}
\text{SQ} = \frac{\sum_{(p,g) \in \text{TP}} \text{IoU}(p, g)}{|\text{TP}|}
\end{equation}
\begin{equation}
\text{RQ} = \frac{|\text{TP}|}{|\text{TP}| + \frac{1}{2}|\text{FP}| + \frac{1}{2}|\text{FN}|}
\end{equation}
For a more comprehensive evaluation, we compute PQ, RQ, and SQ scores separately for the \textit{Things} and \textit{Stuff} classes.

For the image caption task, Bilingual Evaluation Understudy (BLEU) \cite{papineni2002bleu} is used to evaluate the quality of generated texts. Specifically, BLEU-4 evaluates the overlap of n-grams between the generated sentences and the reference text for $n = 1, 2, 3, 4$.

The BLEU score is defined as:
\begin{equation}
\text{BLEU} = BP \cdot \exp \left( \sum_{n=1}^{4} w_n \log p_n \right)
\end{equation}

Where:
\begin{equation}
p_n = \frac{\text{Count of matching n-grams}}{\text{Count of total n-grams in generated text}}
\end{equation}

$w_n$ is the weight for n-grams, typically set to $\frac{1}{4}$ for BLEU-4.

The brevity penalty, $BP$, is defined as:
    \begin{equation}
    BP = 
    \begin{cases} 
        1 & \text{if } c > r \\
        \exp(1 - \frac{r}{c}) & \text{if } c \leq r 
    \end{cases}
    \end{equation}
where $c$ is the length of the generated translation and $r$ is the effective reference corpus length.

The BLEU score ranges from 0 to 1, with 1 indicating a perfect match with the reference text. Before computing the BLEU-4 scores, we used beam search to obtain descriptive sentences. The beam sizes employed in the experiments were 3 and 5.

\subsection{Implementation Details}
All experiments are conducted on two NVIDIA RTX 4090 GPUs. We employ ResNet-50~\cite{he2016deep} pre-trained on ImageNet~\cite{deng2009imagenet} as the backbone for the image encoder in the pixel-level module. The total number of training epochs are 75 for all experiments. The learning rate was initialized as 0.0001 and decay to 0.00001 through multi-step decay. The input image resolution is $800 \times 800$ and the training batchsize is set to 4. The number of learnable queries $N$ in the segmentation module is set to 200. The code implementation is based on MMDetection~\cite{mmdetection}.

 \begin{table*}
 	\centering
 	\small
 	\caption{Comparisons between our proposed panoptic perception method and other interpretation approaches on FineGrip. The best results are marked in bold.}
 	\setlength{\tabcolsep}{1mm}{
 	\begin{tabular}{c|c|ccc|ccc|ccc|cc}
 		\toprule[0.4mm]
 		 \multirow{2}*{Task}   &\multirow{2}*{Method}  & \multicolumn{3}{c}{All}  & \multicolumn{3}{c}{Things}   & \multicolumn{3}{c}{Stuff}   & \multicolumn{2}{c}{BLEU score}         \\
 		&                    & PQ            & SQ            & RQ            & $\text{PQ}^{Th}$            & $\text{SQ}^{Th}$           &
 	$\text{RQ}^{Th}$         & $\text{PQ}^{st}$            & $\text{SQ}^{St}$           & $\text{RQ}^{St}$      & beam size=3   & beam size=5   \\
 		\midrule 
 		 \multirow{3}*{\makecell{Panoptic \\Segmentation}}& Panoptic FPN~\cite{PanopticFPN}     &54.1    &78.6	       &68.3	    &55.8 	&79.0 	&\textbf{70.7}	&47.1	&77.1 	&58.8          & -             & -             \\
 		& MaskFormer \cite{MaskFormer}       &50.2	&79.4 	&63.1	&49.9 	&78.8 	&63.3	&51.6  	&81.8 	&62.5         & -             & -             \\
 		& Mask2Former~\cite{Mask2Former}  &55.2 	&\textbf{81.0}	&67.9	&54.8 	&80.5 	&67.9	&57.0 	&83.0 	&68.0  & -             & -             \\
 		\midrule
 		 \multirow{4}*{ \makecell{Caption \\Generation }}  & SAT~\cite{xu2015show}  & -    & -     & -     & -      & -      & -     & -       & -     & -     & 30.4       & 29.4      \\
 		& SAT w/o attention  & -       & -      & -       & -       & -       & -       & -        & -       & -         & 36.2          & 35.9          \\
 		& MLAT~\cite{9709791} w/o LSTM     & -      & -      & -       & -      & -        & -      & -      & -         & -        & 35.9     & 35.8       \\
 		& MLAT~\cite{9709791}               & -       & -        & -       & -        & -       & -       & -      & -        & -         & 41.2 & 40.3           \\
 		\midrule
 		\multirow{3}*{\makecell{Panoptic \\Perception}}  & Ours w/ SAT       & 49.6          & 79.5       & 62.1          & 48.5          & 78.7          & 61.4          & 53.8          & \textbf{82.6}          & 64.7          & \textbf{45.2}          &\textbf{44.0 }\\
 		& Ours w/ MaskFormer & 50.9          & 79.3          & 63.8          & 49.8          & 78.6          & 63.2          & 55.1          & 82.1          & 66.1          & 42.4          & 41.7          \\
 		& Ours w/ Mask2Former & \textbf{56.5}     &80.9     &\textbf{69.6}   &\textbf{56.3}           &\textbf{80.6}         &69.7           &\textbf{57.3}           &82.2           &\textbf{68.9}           &42.3           &41.5\\
 		\bottomrule[0.4mm]
 	\end{tabular}}     
 	\label{table-comparison}
 \end{table*}
 
\subsection{Quantitative Analysis}

We evaluate the proposed panoptic perception model on multi-task and multi-modal interpretation. As seen in Table~\ref{table-comparison}, we conduct the proposed model with traditional panoptic segmentation and caption generation tasks. For panoptic segmentation, the PQ, SQ, RQ for all classes, \textit{thing} classes and \textit{stuff} classes are computed respectively. While for captioning task, two kinds of beam sizes are reported. 

\textbf{Segmentation performance}. We embed the proposed model-agnostic panoptic perception method into several baselines including MaskFormer~\cite{MaskFormer} and Mask2Former~\cite{Mask2Former}, of which the models support joint optimization with end-to-end training. Compared to the single-task segmentation model, our panoptic model achieves 0.7\% and 1.3\% PQ improvement, respectively. Further we respectively explore the segmentation performance on \textit{thing} and \textit{stuff} classes. On the one hand, ours w/ MaskFormer achieves effective improvement on \textit{stuff} classes. On the other hand, ours w/ Mask2Former achieves solid advancements on \textit{thing} classes. We argue that cross-modal joint training can boost divergent single-task performance.

\textbf{Captioning performance}. Regarding caption generation, we compare two baseline models including SAT~\cite{xu2015show} and MLAT~\cite{9709791}. We evaluate the performance of a variant of SAT without attention mechanisms and a variant of MLAT without LSTM. For fair comparisons, we conduct panoptic perception using both the LSTM decoder (ours w/ SAT) and the Transformer decoder (ours w/ MaskFormer and ours w/ Mask2Former) for the caption module. The quantitative results indicate that joint optimization of the panoptic perception model significantly enhances the performance of caption generation. Concretely, our model with LSTM decoder significantly outperforms original captioning models SAT~\cite{xu2015show} and gains an improvement of 14.8\% (beam-size=3) and 14.6\% (beam-size=5) in BLEU score. While the Transformer-decoder-based panoptic perception also achieves favourable improvements on both segmentation and captioning tasks. 

\textbf{Cross-modal sub-tasks interaction}. We demonstrate that joint optimization across multi-modal tasks within an intact model can advance each single-task performance. As shown in Table~\ref{table-comparison}, the proposed panoptic perception methods support model-agnostic combinations with current segmentation models and achieve pixel-wise classification, instance-wise segmentation and image captioning concurrently. And compared to the single segmentation or captioning task, the panoptic perception achieve higher performance on both tasks. To our knowledge, cross-modal interactive optimization can advance the overall performance. The experimental results further validate this perspective. Particularly, ours w/ Mask2Former achieves 1.3\% and 1.7\% improvement in PQ and RQ over all classes to the original Mask2Former. While in the fine-grained \textit{things} classes, the metrics are also significantly improved by 1.5\% and 1.8\&, respectively.

\begin{figure*}[htbp]
    \centering
    \includegraphics[scale=0.4]{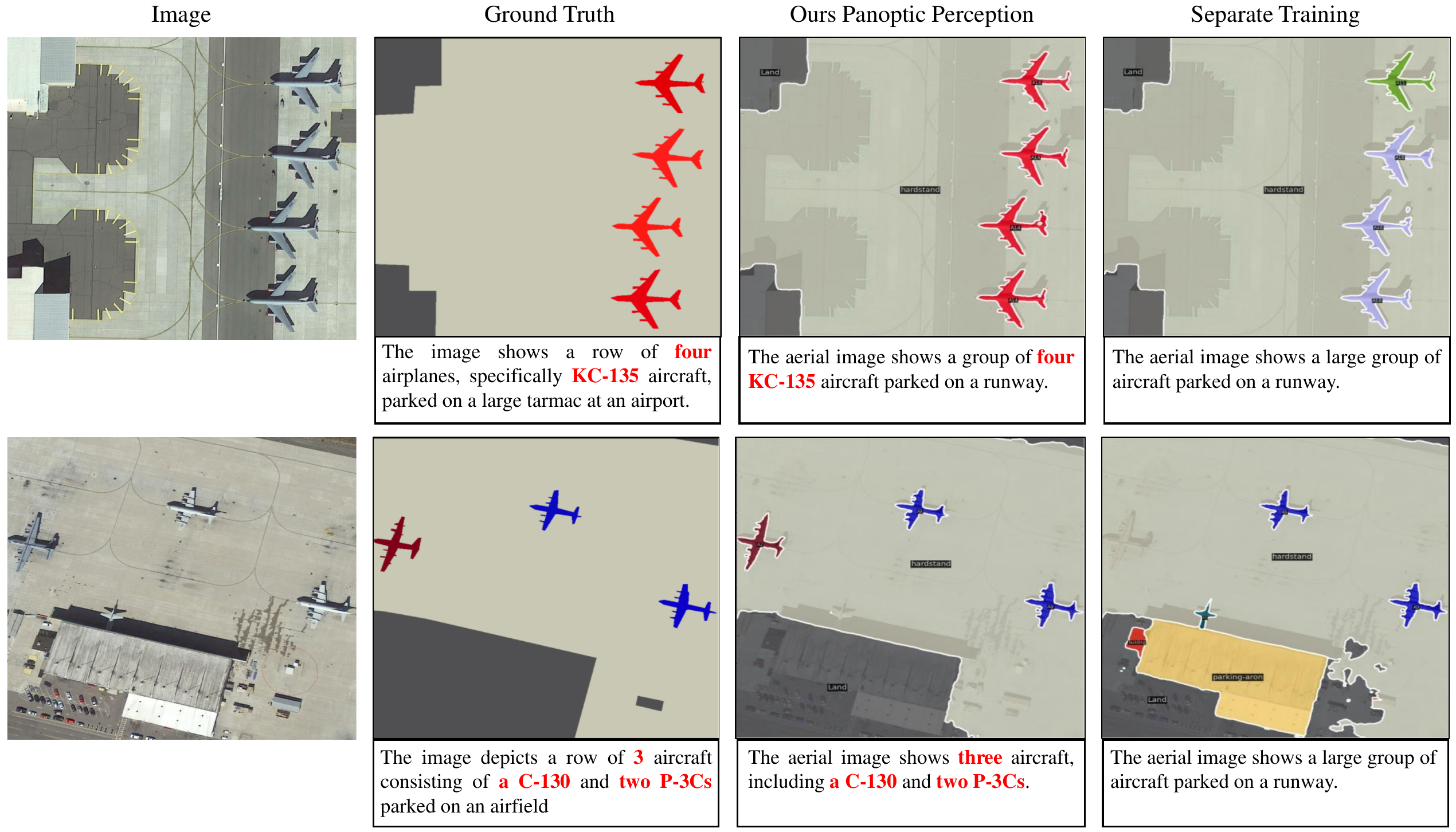} 
    \caption{Panoptic perception results from the jointly trained and separately trained models. The first column displays the ground-truth mask and one of the five caption sentences for each image.}
    \label{fig-vis_comparison} 
\end{figure*}

Fig.~\ref{fig-vis_comparison} illustrates the qualitative comparisons of the panoptic perception results with separate training approach. In the first row, our panoptic perception model can accurately perceive the categories and quantities of the foreground airplane targets in both segmentation and description branches. While in the second row, the foreground and background semantics are both recognized with fine-grained annotations. In contrast, the predictions of the separately trained model are coarser in both segmentation and captioning tasks. Particularly in the captioning task, the model trained independently struggles to generate accurate fine-grained captions due to the lack of additional information from the segmentation task. 

\begin{figure*}[htbp]
    \centering
    \includegraphics[scale=0.48]{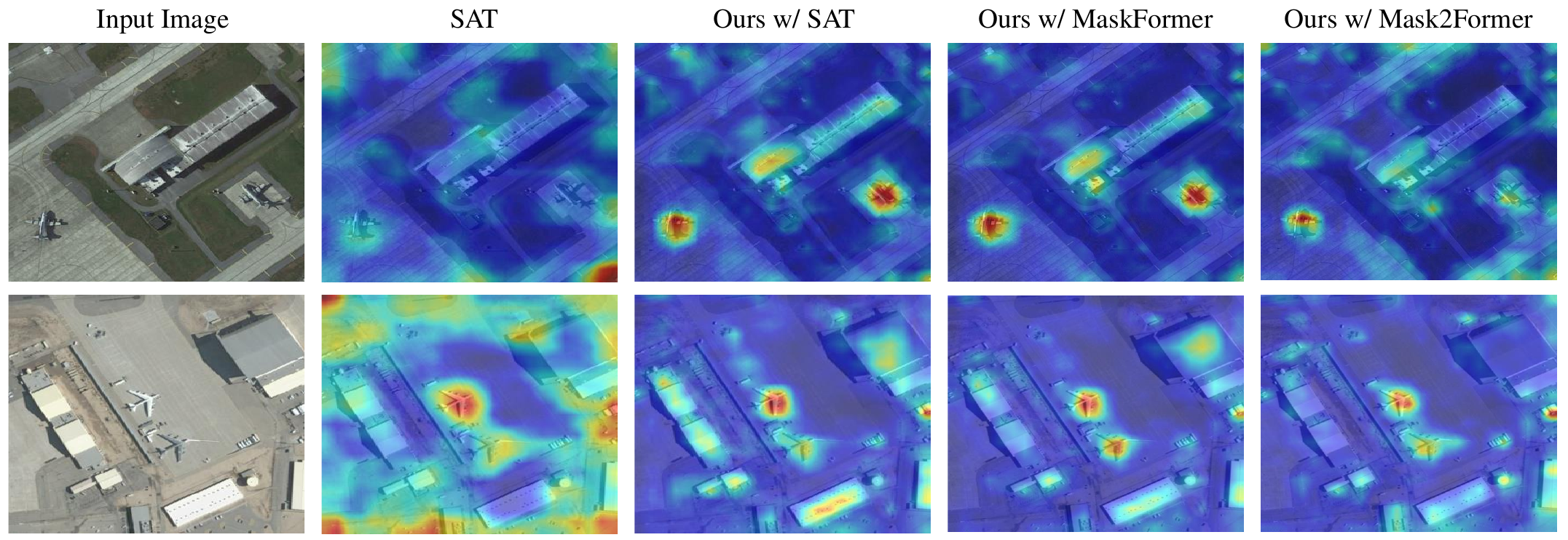} 
    \caption{Saliency maps of features from the last layer of the ResNet-50 backbone for different models. The regions with colors closer to red indicate areas of higher feature response intensity.}
    \label{fig-feature_map_vis} 
\end{figure*}

To intuitively demonstrate the impact of joint optimization on model behavior, we visualize the feature saliency maps for SAT and our panoptic perception model w/ SAT, w/ MaskFormer and w/ Mask2Former, respectively. We select the output feature map of the last layer of the ResNet-50 backbone with a shape of $2048\times25\times25$. After being averaged over channels and normalized, these feature maps are overlaid onto the original input images. As illustrated in Fig.~\ref{fig-feature_map_vis}, the SAT model pays more attention to the overall image while overlooking the foreground semantics. This is because the caption generation task requires the model to generalize over the entire image, which prevents the model from focusing on specific targets. On the contrary, ours w/ SAT focus on both foreground and background semantics. The Transformer decoder based panoptic perception methods also strike a balance by emphasizing foreground instances and maintaining a focus on the overall image. This validates the effectiveness of a unified multi-task framework and joint optimization strategy in enhancing a more comprehensive interpretation of RSIs.

\subsection{Ablation Study}
\label{Sec:Abla}
\textbf{Collaborative Optimization}. The proposed joint optimization-based panoptic perception model leverages both the segmentation branch and captioning branch with task-interactive optimization. Thus the weight contribution within multi-modal tasks has a vital impact on the joint optimization and multi-task learning problem. In Eqn.~\ref{eqn-lambda}, $\lambda$ was used to balance the contribution of multi-modal branches during training. As shown in Table~\ref{table-abla-lambda}, we respectively compute the segmentation performance and the captioning score. The weight to balance the contribution of sub-tasks is not a fixed value. Concretely, regarding MaskFormer as the baseline, it achieves the highest 50.9\% PQ when $\lambda=2.0$, while the best BLEU score arrives with $\lambda=0.7$. We infer that a lower $\lambda$ prevents the caption module from providing sufficient instance-level information, while a higher $\lambda$ can overwhelm the impact of the few instance-related words with the major non-instance parts of the sentence. While ours w/ Mask2Former achieves the highest PQ at $\lambda=1.0$. To our knowledge, the gradient contributions from different sub-tasks should maintain balanced or tilted towards vital tasks, which have been investigated in multi-task learning problems.

\textbf{Captioning Decoder}. In our proposed panoptic perception model, we develop a universal Transformer decoder for captioning task, which is parallel and interactive with the Transformer decoder in the segmentation branch. Table~\ref{table-abla-Nlayer} displays the panoptic perception results of our model with a Transformer decoder at different layer counts. Regarding Mask2Former as the base model, We observe that segmentation performance improves with increased model complexity, reaching 56.5\% PQ with three decoder layers. However, The caption module performs best when applying two decoder layers. It achieves a 43.4\% BLEU score at a beam size of 3, surpassing the model based on the SAT decoder. 

There are two reasons for the significant difference in the impact of decoder layer counts on segmentation and captioning performance. On the one hand, substantial model complexity presents challenges in achieving efficient optimization, thereby hindering the caption performance. On the other hand, a more complex structure possesses stronger feature representation capabilities. It can encourage the backbone to learn feature representations containing richer information, ultimately leading to improved performance in the segmentation branch.

 \begin{table}[tbp]
 	\caption{The impact of $\lambda$. }
 	\centering
 	\setlength{\tabcolsep}{2mm}{
 		\begin{tabular}{l|c|ccc|cc}
 			\toprule[0.4mm]
			\multirow{2}*{Model} &\multirow{2}*{$\lambda$}&\multicolumn{3}{c}{All}&\multicolumn{2}{c}{BLEU score}\\
			&&PQ&SQ&RQ&bs=3&bs=5\\
			\midrule
			\multirow{5}*{\makecell{Ours w/\\MaskFormer}}&0.5 &49.4	&\textbf{79.5}	&61.9	&43.2	&42.1\\
			&0.7 &50.1	&79.4	&62.9	&\textbf{43.8}	&\textbf{43.2}\\
			&1.0 &50.2	&79.2	&63.1	&43.1	&42.6\\
			&2.0 &\textbf{50.9}	&79.3	&\textbf{63.8}	&42.4	&41.7\\
			&3.0 &50.7	&79.4	&63.5	&42.6 &42.2\\
			\midrule
			\multirow{5}*{\makecell{Ours w/\\Mask2Former}}&0.5 &54.0	&81.0	 &66.3 	&41.8	&40.8\\
			&0.7 & 54.4	&\textbf{81.1}	&66.6	&41.3	&40.7 \\
			&1.0 &\textbf{56.5} &80.9	&\textbf{69.6}	&42.3	&\textbf{41.5} \\
			&2.0 & 54.1	&81.0   &66.6	&40.4	&40.8 \\
			&3.0 & 54.0	&81.0	&66.4	&\textbf{42.8}	&41.4 \\
 			\bottomrule[0.4mm]
 	\end{tabular}}
 	\label{table-abla-lambda}
 \end{table}

 \begin{table}[tbp]
 	\caption{The impact of the decoder layer number $N_{cap}^{dec}$ in the caption module.}
 	\centering
 	\setlength{\tabcolsep}{2mm}{
 		\begin{tabular}{l|c|ccc|cc}
 			\toprule[0.4mm]
			\multirow{2}*{Model} &\multirow{2}*{$N_{cap}^{dec}$}&\multicolumn{3}{c}{All}&\multicolumn{2}{c}{BLEU score}\\
			&&PQ&SQ&RQ&bs=3&bs=5\\
			\midrule
			\multirow{4}*{\makecell{Ours w/\\MaskFormer}}&6&50.5 	&79.1 	&63.6	&41.6	&41.5\\
			&3&\textbf{50.9}	&79.3	&\textbf{63.8}	&42.4	&41.7\\
			&2&50.6	&\textbf{79.4} 	&63.5	&\textbf{42.7}	&\textbf{42.6}\\
			&1&49.3	&79.3	&62.0	&42.1	&36.5\\
			\midrule
			\multirow{4}*{\makecell{Oursw w/\\Mask2Former}}&6 &53.8	&\textbf{81.2}	&66.1	&41.3	&40.8	\\
			&3&\textbf{56.5}	&80.9	&\textbf{69.6}	&42.3	&41.5	\\
			&2&53.4 &80.9 	&65.7	&\textbf{43.4}	&\textbf{42.7}	\\
			&1&54.6 	&81.1 	&67.2	&42.8	&42.3\\
 			\bottomrule[0.4mm]
 	\end{tabular}}
 	\label{table-abla-Nlayer}
 \end{table}

\textbf{Model Attributes}. We evaluate the proposed panoptic perception model attributes from parameters, FLOPs and inference FPS. The results are reported in Table~\ref{table-abla-param}. 
In addition, we also analyze the category-wise improvements of the proposed panoptic perception model to the base segmentation model. Table~\ref{table-abla-classwise} shows the gain on PQ of the proposed model, especially on the challenging fine-grained foreground semantics, proving the effectiveness of boosting single-task interpretation.

 \begin{table}[t]
 	\caption{The parameters, FLOPs and inference FPS of the proposed panoptic perception models. }
 	\centering
 	\setlength{\tabcolsep}{2mm}{
 		\begin{tabular}{l|ccc}
 			\toprule[0.4mm]
 			Model &Param. (M)&FLOPs (G)&FPS   \\
 			\midrule
 			MaskFormer &45.0&126&8.5\\
 			Ours w/ MaskFormer &52.1&135&4.8\\
 			\midrule
 			Mask2Former &44.0 &160 &6.8\\
 			Ours w/ Mask2Former &46.5&172&6.2\\
 			\bottomrule[0.4mm]
 	\end{tabular}}
 	\label{table-abla-param}
 \end{table}

 \begin{table*}[htbp]
 	\scriptsize
 	\caption{Quantitative comparison in category-wise PQ (\%).}
 	\centering
 	\setlength{\tabcolsep}{0.5mm}{
 			\begin{tabular}{l|c|c|c|c|c|c|c|c|c|c|c|c|c|c|c|c|c|c|c|c|c|c|c|c|c}
 					\toprule[0.4mm]
 					Method &A1&A2&A3&A4&A5&A6&A7&A8&A9&A10&A11&A12&A13&A14&A15&A16&A17&A18&A19&A20&Land&Run.&Hards.&Park.&Buil.\\
 					\midrule
					Mask2Former &45.8&34.3&62.0&65.2&46.1&75.5&71.5&55.9&53.4&75.0&60.8&74.9&31.8&39.6&43.1&27.7&73.6&45.5&54.3&59.1&82.0&36.4&81.7&34.0&50.8\\
					Ours w/ Mask2Former &46.0&38.3&63.7&67.9&49.5&75.7&69.4&55.9&54.6&75.1&59.9&75.1&39.3&39.0&49.9&30.9&73.3&45.3&52.6&64.2&81.7&35.9&82.5&35.4&51.2 \\
					gain&+0.2&+4.0&+1.7&+2.7&+3.4&+0.2&-2.1&0.0&+1.2&+0.1&-0.9&+0.2&+7.5&-0.6&+6.8&+3.2&-0.3&-0.2&-1.7&+6.1&-0.3&-0.5&+0.8&+1.4&+0.4\\
 					\bottomrule[0.4mm]
 			\end{tabular}}
 	\label{table-abla-classwise}
 \end{table*}

\section{Conclusion} 
In this paper, we develop a novel fine-grained panoptic perception task to achieve a more comprehensive unified interpretation of RSIs. The main contributions include: 1) We introduce FineGrip, a novel fine-grained panoptic perception dataset. 2) We propose an end-to-end model-agnostic panoptic perception method that integrates multi-task joint optimization covering pixel-level classification, instance-level segmentation and image-level captioning within an intact model. 3) An efficient semi-automatic annotation system that provides scalability for segmentation annotations. Experimental results demonstrate the effectiveness of the proposed unified architecture for panoptic perception. Particularly, the feasibility of joint optimization across multiple tasks by interactive learning was validated. However, the consistency across multiple sub-tasks is still a challenging but promising research priority. Our future work will further expand the proposed dataset and improve the task-interactive facilitation of the panoptic perception model.


%

\ifCLASSOPTIONcaptionsoff
  \newpage
\fi



%
\bibliographystyle{IEEEtran}
\bibliography{refs}




\end{document}